%% file: acl_latex.tex
\documentclass[11pt]{article}

\usepackage[preprint]{acl}
\usepackage{amsmath}
\usepackage{times}
\usepackage{latexsym}
\usepackage{amsfonts,amssymb}
\usepackage{pifont}
\newcommand{\cmark}{\ding{51}} 
\newcommand{\xmark}{\ding{55}} 
\usepackage{newunicodechar}    
\newunicodechar{✅}{\cmark}
\newunicodechar{❌}{\xmark}
\usepackage{multirow} 
\usepackage[T1]{fontenc}
\newcommand{\blue}[1]{#1}
\usepackage{xcolor}
\usepackage{tcolorbox}
\usepackage{booktabs}   
\usepackage{array}      
\usepackage{tabularx}   
\newcolumntype{Y}{>{\raggedright\arraybackslash}X} 
\tcbuselibrary{skins,breakable,listings}
\definecolor{chemMintBack}{HTML}{F2FBF7}
\definecolor{chemMintBar}{HTML}{E3F6EE}
\definecolor{chemMintFrame}{HTML}{7DC9AE}
\definecolor{chemLavBack}{HTML}{F5F2FF}
\definecolor{chemLavBar}{HTML}{E9E4FF}
\definecolor{chemLavFrame}{HTML}{A89BE8}
\definecolor{chemPeachBack}{HTML}{FFF6EF}
\definecolor{chemPeachBar}{HTML}{FFE6D6}
\definecolor{chemPeachFrame}{HTML}{F2B091}
\usepackage[dvipsnames]{xcolor} 
\definecolor{chemLabelBlue}{HTML}{1F6FEB} 
\definecolor{chemLabelbrown}{HTML}{8B4513} 
\definecolor{fieldlabelbrom}{HTML}{A0522D} 

\newcommand{\chempromptcolorsMint}{%
  \colorlet{chemBack}{chemMintBack}%
  \colorlet{chemTitle}{chemMintBar}%
  \colorlet{chemFrame}{chemMintFrame}%
}

\chempromptcolorsMint
\makeatletter
\lstdefinestyle{chemPromptStyle}{
  basicstyle=\ttfamily\footnotesize,
  breaklines=true,
  columns=fullflexible,
  literate=
    {System:}{{{\color{chemLabelbrown}\bfseries System:}}}{7}
    {User:}{{{\color{chemLabelbrown}\bfseries User:}}}{5}
    {Assistant:}{{{\color{chemLabelBlue}\bfseries Assistant:}}}{10}
    {EXAMPLE:}{{{\color{chemLabelBlue}\bfseries EXAMPLE of Atomic Priors Knowledge Base:}}}{8}
    {Schema:}{{{\color{chemLabelBlue}\bfseries Fields Meaning:}}}{7}
    {SMILES:}{{{\color{chemLabelBlue}\bfseries SMILES:}}}{6}
    {GT:}{{{\color{chemLabelBlue}\bfseries Ground Truth:}}}{3}
    {Response:}{{{\color{chemLabelBlue}\bfseries Response:}}}{9}
    {<analysis>}{{{\color{chemLabelBlue}\bfseries <analysis>}}}{10}
    {</analysis>}{{{\color{chemLabelBlue}\bfseries </analysis>}}}{11}
    {<answer>}{{{\color{chemLabelBlue}\bfseries <answer>}}}{8}
    {</answer>}{{{\color{chemLabelBlue}\bfseries </answer>}}}{9}
    {<confidence>}{{{\color{chemLabelBlue}\bfseries <confidence>}}}{12}
    {</confidence>}{{{\color{chemLabelBlue}\bfseries </confidence>}}}{13}
    {Similarity:}{{{\color{chemLabelBlue}\bfseries Similarity:}}}{11}
    {DTS:}{{{\color{chemLabelBlue}\bfseries Discrete Token Sequence:}}}{4}
    {SAF:}{{{\color{chemLabelBlue}\bfseries Selected Atom-Level Features:}}}{4}
    {SMF}{{{\color{chemLabelBlue}\bfseries Selected Molecule-Level Features:}}}{4}
    {IMPORTANT:}{{{\color{chemLabelBlue}\bfseries IMPORTANT:}}}{10}
    {Task:}{{{\color{chemLabelBlue}\bfseries Task:}}}{5}
    {FD:}{{{\color{chemLabelBlue}\bfseries Feature Descriptions:}}}{3}
    {ISA:}{{{\color{chemLabelBlue}\bfseries Instance-Specific Analysis:}}}{4}
}

\makeatother
\newtcblisting{chemprompt}[2][]{%
  enhanced, breakable, listing only, listing engine=listings,
  width=\textwidth,              
  colback=chemBack, colframe=chemFrame,
  boxed title style={colback=chemTitle, colframe=chemFrame},
  title={#2}, fonttitle=\bfseries,
  boxrule=0.6pt, arc=2mm, outer arc=2mm,
  left=1.2mm, right=1.2mm, top=1.2mm, bottom=1.2mm,
  listing options={style=chemPromptStyle},
  #1
}

\usepackage[utf8]{inputenc}

\usepackage{microtype}

\usepackage{inconsolata}

\usepackage{graphicx}

\title{ChemATP: A Training-Free Chemical Reasoning Framework for Large Language Models}

\author{
  Mingxu Zhang\textsuperscript{1},
  Dazhong Shen\textsuperscript{2},
  Qi Zhang\textsuperscript{3},
  Ying Sun\textsuperscript{1}\thanks{Corresponding author.} \\
  \textsuperscript{1}The Hong Kong University of Science and Technology (Guangzhou) \\
  \textsuperscript{2}Nanjing University of Aeronautics and Astronautics \\
  \textsuperscript{3}Shanghai AI Laboratory \\
  \texttt{mzhang630@connect.hkust-gz.edu.cn}, 
  \texttt{shendazhong@nuaa.edu.cn}, \\
  \texttt{zhangqi.fqz@gmail.com},
  \texttt{sunyinggilly@gmail.com}
}

\begin{document}
\maketitle
\begin{abstract}
Large Language Models (LLMs) exhibit strong general reasoning but struggle in molecular science due to the lack of explicit chemical priors in standard string representations.
\blue{Current solutions face a fundamental dilemma. Training-based methods inject priors into parameters, but this static coupling hinders rapid knowledge updates and often compromises the model's general reasoning capabilities. Conversely, existing training-free methods avoid these issues but rely on surface-level prompting, failing to provide the fine-grained atom-level priors essential for precise chemical reasoning.}
To address this issue, we introduce \textbf{ChemATP}, a framework that decouples chemical knowledge from the reasoning engine. By constructing the first atom-level textual knowledge base, ChemATP enables frozen LLMs to explicitly retrieve and reason over this information dynamically.
This architecture ensures interpretability and adaptability while preserving the LLM's intrinsic general intelligence.
Experiments show that ChemATP significantly outperforms training-free baselines and rivals state-of-the-art training-based models, demonstrating that explicit prior injection is a competitive alternative to implicit parameter updates.
\end{abstract}

\section{Introduction}

Large language models (LLMs) exhibit strong reasoning abilities and have driven rapid progress in domains such as automated theorem proving~\citep{jiang2022draft,wang2023lego} and code generation~\citep{AlphaCode, codellama}. However, applying LLMs to molecular science (e.g., molecular property prediction) remains challenging. Current LLM-for-chemistry pipelines typically expose molecules as 1D strings---most commonly SMILES~\citep{weininger1988smiles} or SELFIES~\citep{selfies}---and feed these sequences directly into a text-only backbone. These string encodings lack explicit, fine-grained chemical priors (e.g., per-atom charge, aromaticity, or local topology), which are key information for chemists to reason about reactivity, stability, and molecular properties. As a result, an LLM that only sees SMILES or SELFIES is pushed toward ungrounded, pattern-matching behavior on surface text rather than chemically grounded reasoning, which caps performance and undermines traceability and trust. 

\blue{Broadly, existing efforts fall into two categories, revealing a clear gap. 
Training-based methods incorporate chemical priors by learning additional interfaces that fuse necessary information into LLMs~\citep{cao2023instructmol,liu2023molca,zhang2024unimot,momus}. 
While effective, this paradigm implicitly bakes knowledge into static model parameters. 
This results in inherent rigidity where updating knowledge requires retraining and risks catastrophic forgetting of the LLM's general reasoning capabilities. Furthermore, the reliance on opaque weight updates undermines the traceability essential for scientific discovery.
Conversely, training-free approaches keep LLMs frozen and rely on prompting or external molecular descriptors~\citep{xian2025molrag,llm4sd,chen2025reactgpt,zhang-etal-2025-automated}. 
However, these methods typically expose only coarse string- or molecule-level information, or delegate prediction to simple downstream models, thus failing to provide the fine-grained, atom-level priors necessary for precise end-to-end reasoning.}

\blue{To address this dilemma, we propose a paradigm shift: decoupling domain-specific chemical knowledge from the general reasoning engine. 
We ask: can we equip frozen LLMs with explicit, fine-grained chemical priors through an external knowledge module, thereby avoiding the rigidity and opacity of parametric updates? 
To this end, we introduce ChemATP (\textbf{Chem}ical \textbf{A}tomic \textbf{T}extualized \textbf{P}riors), a framework that enables LLMs to perform chemically grounded reasoning by retrieving and operating on an atom-level textual knowledge base. 
Unlike training-based methods that compress priors into fixed weights, ChemATP externalizes knowledge: it records atomic attributes (e.g., charge, aromaticity) and molecule-level descriptors in a human-readable format. 
Our design goes beyond simple evidence lookup; it utilizes retrieved structured exemplars to teach the LLM how to interpret these priors for the target task. 
Concretely, ChemATP follows a three-stage pipeline: (i) it parses the task and autonomously selects relevant prior fields, (ii) it retrieves top-$k$ similar molecules to ground the query, and (iii) it performs integrative reasoning over the explicit atom-level fields to produce a traceable decision. 
Experiments show that by leveraging explicit priors rather than implicit parameters, ChemATP not only achieves state-of-the-art performance among training-free baselines but is also competitive with, or even surpasses, specialized training-based methods.}

In summary, our contributions are threefold. First, we construct the first atom-level textual knowledge base, which externalizes fine-grained chemical priors  to overcome the information bottleneck of standard linear strings. 
Second, we propose ChemATP, a framework that decouples knowledge from the reasoning engine, enabling frozen LLMs to perform interpretable, dynamically updateable inference by explicitly consulting these priors rather than relying on static parametric memory. 
Finally, we provide strong empirical evidence demonstrating that this explicit grounding paradigm not only consistently outperforms training-free baselines but also rivals or exceeds training-based methods, validating that externalized knowledge is a competitive alternative to implicit weight updates for scientific reasoning.

\begin{figure*}[t]
    \centering
    \includegraphics[width=\linewidth]{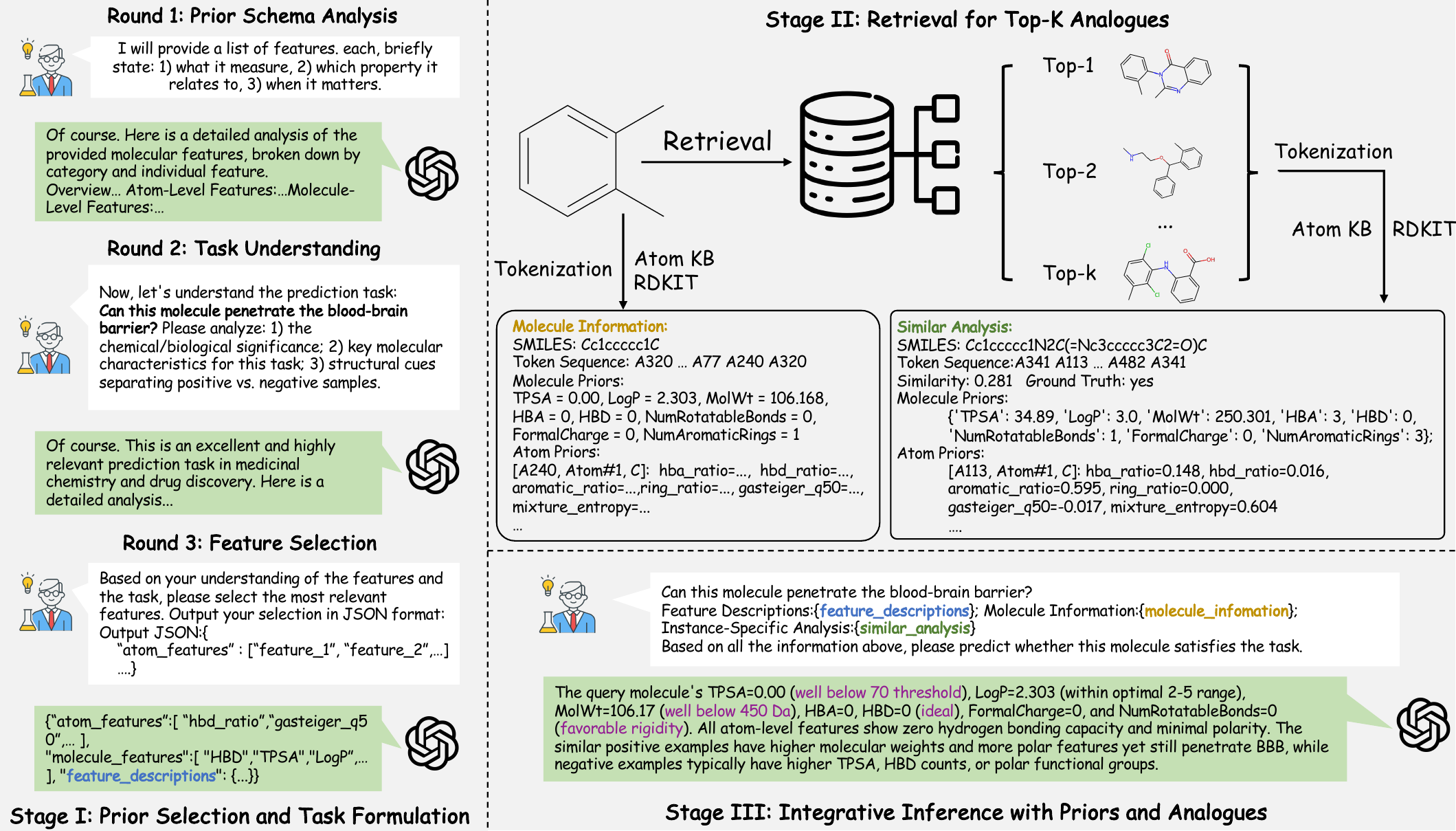}
    \caption{Overview of ChemATP. ChemATP is a training-free, three-stage framework in which an LLM (i) analyzes the prior schema and selects task-relevant atom- and molecule-level priors, (ii) retrieves top-$k$ analogue molecules via Morgan fingerprint similarity and attaches their priors and labels as exemplars, and (iii) integrates the query molecule’s priors, the retrieved analogues, and the feature description in a structured prompt to make the inference.}
    \label{fig:1}
\end{figure*}

\section{Related Work}

\paragraph{Molecule Representation Learning.}
Molecular representations have evolved from early 2D graph encoders to increasingly expressive geometric and multi-modal models. Graph neural networks form the foundation of many 2D approaches and are often enhanced with self-supervised objectives such as mutual-information maximization, attribute masking, and context prediction~\citep{kipf2016semi,velivckovic2017graph,xu2018powerful,velivckovic2018deep,sun2019infograph,you2020graph,inae2023motif,hu2019strategies}. More recently, 3D-aware and equivariant architectures have been introduced to directly model molecular geometry
~\citep{schutt2017schnet,gasteiger2020directional,fuchs2020se,zhou2023uni,alcaide2024uni,stark20223d}. To bridge 2D and 3D representations, GraphMVP leverages geometric views to supervise 2D pretraining through contrastive learning~\citep{liu2022pretraining}. Beyond geometry, multi-modal methods align molecular structures with textual descriptions to learn transferable embeddings~\citep{liu2023multi}. At a finer granularity, FragmentNet masks chemically meaningful fragments during pretraining~\citep{samanta2025fragmentnet}, while AtomDisc~\citep{zhang2025atom} introduces the first atom-level tokenizer that maps each atom to a discrete token, enabling more detailed molecular understanding. \blue{ICRL~\citep{ICRL} pioneers latent in-context learning by constructing in-context examples with embeddings from a domain-specific foundation model, showing that domain embeddings can serve as an alternative interface for induction and inference.}

\paragraph{Chemical Large Language Models.}
Recent work has sought to inject chemical understanding into LLMs through training or adaptation. One approach is domain-specific pre-training: for example, ChemDFM~\citep{zhao2024chemdfm} was trained on 34 billion tokens of chemical papers and textbooks, producing a 13B-parameter model that can converse in chemistry and even outperform GPT-4 on many benchmark tasks. Another approach is multi-modal integration, aligning molecular structure representations with text. MolXPT~\citep{liu2023molxpt} wraps SMILES sequences with natural language context during pre-training, effectively teaching the model to understand molecular strings in a linguistic manner. InstructMol~\citep{cao2023instructmol} takes a two-stage instruction-tuning approach, combining graph-based features with text so that a language model can follow chemical instructions and questions reliably. Beyond these training-intensive methods, some works expand the LLM input/output space to handle chemical structures explicitly like UniMoT~\citep{zhang2024unimot} and AtomDisc~\citep{zhang2025atom}, discrete chemical priors into tokens and append them into vocabulary.

\paragraph{Retrieval Augmented Generation in Chemistry.}
MolRAG~\citep{xian2025molrag} introduces a retrieval-augmented generation framework which retrieves a few analogues of a query molecule from a database and inserts them into a prompted chain-of-thought, allowing the LLM to reason with concrete structural examples. f-RAG~\citep{lee2024molecule} augments a pretrained generator with fragment retrieval to steer synthesis toward valid and novel compounds while balancing exploration–exploitation. ChatDrug~\citep{liu2024conversational} combines retrieval with domain feedback in a conversational loop, achieving SOTA across a broad suite of editing tasks. Tool-augmented chemistry agents such as ChemCrow~\citep{m2024augmenting} integrate LLMs with dozens of domain tools and databases, effectively using retrieval to plan multi-step synthesis and answer complex queries. In contrast to LLM4SD~\citep{llm4sd} approaches that handed off decisions to downstream models, these RAG frameworks keep the LLM in the driver’s seat – the model actively consults a knowledge base or examples and then reasons end-to-end.

\section{Method}
ChemATP implements a decoupled paradigm where reasoning is separated from knowledge storage. 
The primary challenge, however, is the absence of existing textual knowledge base capable of providing fine-grained atomic priors for LLMs. To address this gap, we first construct an atom-level textual base and then design a three-stage pipeline (shown in Fig.~\ref{fig:1}) to utilize it: 
\textbf{Stage I} selects task-relevant prior fields; 
\textbf{Stage II} retrieves top-$k$ similar molecules as exemplars; 
and \textbf{Stage III} integrates these explicit priors to generate the final decision.

\subsection{Atomic Textualized Prior Knowledge Base}
\label{sec:construct_kb}
\paragraph{Atom Tokenization.}
We use the AtomDisc tokenizer~\citep{zhang2025atom} without modification. 
In AtomDisc, MoleculeSTM~\citep{liu2023multi} maps each atom $v$ in a molecular graph $G=(V,E)$ to an embedding $\mathbf{r}_v\in \mathbb{R}^d$. 
A vector-quantized codebook $\mathcal{C}=\{\mathbf{c}_j\}_{j=1}^{K}$ is trained offline; each atom is assigned the index of its nearest codeword
\[
\ell_v=\arg\min_{1\le j\le K}\|\mathbf{r}_v-\mathbf{c}_j\|_2^2,
\]
and the discrete token is $A_{\ell_v}$. 
We directly use these fixed AtomDisc tokens to form the atom-token sequence $\mathbf{A}(G)=(A_{\ell_{v_1}},\dots,A_{\ell_{v_{|V|}}})$ as inputs to our textualized prior knowledge base.

\paragraph{Knowledge Base Construction.}
We construct the atomic textualized prior knowledge base using a large corpus of molecules sampled from PubChem~\citep{pubchem}. 
For each molecule, we first derive its AtomDisc atom-token sequence $\mathbf{A}(G)$. 
We then compute a comprehensive set of RDKit-based descriptors for every atom, covering topology (e.g., ring environment, aromaticity, hybridization), electronic properties (e.g., formal and Gasteiger charges), hydrogen-bonding roles, and local chemical context (e.g., neighbor types and bond attributes). 
To generate a unified profile for each atom token $t$, we aggregate all corresponding instances from the corpus. 
Numeric attributes are summarized using robust statistics, specifically the median and interquartile range (IQR), while categorical and boolean attributes are described by their modes and empirical distributions. 
Furthermore, to capture typical local chemical environments, we compute the Pointwise Mutual Information (PMI) between token $t$ and its 1-hop neighbors $t'$: $\mathrm{PMI}(t,t')=\log\frac{P(t,t')}{P(t)\,P(t')}$. We retain neighbors with the highest PMI scores. Finally, each atom profile is materialized as a structured JSON record to support retrieval.

\subsection{Stage I: Prior Selection and Task Formulation}
\label{sec:stage_1}

To provide global context and stabilize reasoning, beyond atom-level priors we also incorporate molecule-level priors computed with RDKit~\citep{rdkit} (e.g., TPSA, MolWt; full list in Appendix~\ref{appendix:kb_schema}). As shown in Fig.~\ref{fig:1}, we implement this stage as a multi-turn dialogue to guide the LLM through a step-by-step reasoning process:
\begin{enumerate}
    \item \textbf{Prior schema analysis.} In the initial turn, the LLM is prompted to interpret the definitions of available priors, summarizing what each schema measures and identifying scenarios where it provides informative signals.
    \item \textbf{Task-aware reasoning.} Building on the analysis from the previous turn, the LLM analyzes the specific task instruction and identifies discriminative structural cues or properties.
    \item \textbf{Autonomous feature selection.} In the final turn, the LLM determines the optimal subset of priors for the task. It outputs a machine-readable \texttt{JSON} file listing the selected prior fields, along with their explanations to serve as the reasoning basis for subsequent stages.
\end{enumerate}
The full prompt template for Stage~I is provided in Appendix~\ref{appendix:prompt_example_stage1}.

\subsection{Stage II: Retrieval of Top-k Analogues}
\label{sec:stage_2}

To ground the model's reasoning in concrete chemical evidence, we implement an analogue retrieval mechanism based on structural similarity. 
We precompute Morgan fingerprints for the training corpus and calculate the Tanimoto similarity between the query molecule and potential candidates:
$$
    s(\mathbf{a},\mathbf{b})=\frac{\mathbf{a}\cdot\mathbf{b}}{\|\mathbf{a}\|_1+\|\mathbf{b}\|_1-\mathbf{a}\cdot\mathbf{b}}.
$$
The top-$k$ molecules with the highest similarity scores are selected as in-context exemplars.

For each retrieved analogue, we construct a comprehensive evidence packet containing: (i) the basic SMILES and AtomDisc token sequences; (ii) the task-specific priors selected in Stage I; and (iii) the ground truth label and similarity score. 
However, directly including full atomic profiles for all analogues would introduce excessive noise and dilute the model's attention. 
To address this, we apply a functional atom filtering strategy. Instead of retaining all atoms, we selectively preserve only functional atoms, specifically heteroatoms, charged centers, and key carbons involved in aromaticity or hydrogen bonding while pruning generic backbone atoms (detailed filtering rules in Appendix~\ref{appendix:atom_card_selection}). 
These optimized exemplars effectively provide the LLM with focused, high-density chemical contexts for the subsequent reasoning stage.

\input{results/classification}

\subsection{Stage III: Integrative Inference with Priors and Analogues}
\label{sec:stage_3}
Stage III executes the final decision-making process by synthesizing two complementary knowledge streams: 
(i) the task-specific reasoning evidence distilled in Stage I, and 
(ii) the empirical evidence retrieved in Stage II. 
Rather than relying on opaque parametric memories, we construct a composite reasoning context that presents the LLM with a unified view of chemical principles and instance-specific data. This enables the model to perform explicit, interpretable inference through the following mechanisms:

\paragraph{Explicit priors as inductive bias.}
The feature subset selected in Stage I acts as an explicit inductive bias for the reasoning process. 
By defining which atomic and molecular attributes (e.g., electronegativity, aromaticity) are diagnostically relevant, these priors guide the LLM to attend to chemically causal mechanisms rather than spurious structural correlations. This ensures that the reasoning is grounded in domain principles.

\paragraph{Analogues as anchors.}
The top-$k$ analogues retrieved in Stage II serve as empirical anchors. 
They illustrate how the selected priors manifest in actual chemical space and correlate with specific labels. 
By comparing the query molecule against these anchors, the LLM can perform reasoning by analogy, aligning conserved features, interpolating between known behaviors, and identifying potential deviations based on local atomic differences.

Finally, we integrate these components into a unified prompt (see Appendix~\ref{appendix:prompt_example_stage3}). This design ensures that the LLM's predictions are both theoretically grounded and empirically supported.


\section{Experiments}
\label{sec:exp}
We evaluate ChemATP on standard molecular property benchmarks to answer four research questions (RQs).
\textbf{RQ1}: Does ChemATP significantly outperform existing training-free baselines, and can it rival the performance of state-of-the-art training-based models?
\textbf{RQ2}: What are the individual contributions of atom-level and molecule-level priors? Are they complementary distinct sources of information essential for optimal performance?
\textbf{RQ3}: What is the impact of retrieved analogues on performance, and does ChemATP maintain robustness in zero-shot settings? 
\textbf{RQ4}: Is ChemATP a model-agnostic framework that consistently enhances performance across different LLM backbones?

We address \textbf{RQ1} in Section~\ref{sec:main_results}; \textbf{RQ2} via component analysis in Section~\ref{sec:abla_on_atomic_priors}; \textbf{RQ3} via retrieval ablation in Section~\ref{sec:abla_on_top_k}; and \textbf{RQ4} in Section~\ref{sec:diff_llms}.

\subsection{Experimental Setup}
We evaluate ChemATP on MoleculeNet~\citep{wu2018moleculenet} classification benchmarks and on widely used physicochemical/pharmacokinetic datasets: ESOL~\citep{delaney2004esol,wu2018moleculenet}, FreeSolv~\citep{mobley2014freesolv,wu2018moleculenet}, Lipophilicity (Lipo)~\citep{wu2018moleculenet}, Caco2 permeability~\citep{wang2016adme}, LD50~\citep{zhu2009quantitative}, and AqSol~\citep{sorkun2019aqsoldb}. 
These tasks span solubility (ESOL, FreeSolv, AqSol), lipophilicity (Lipo), absorption (Caco2), and toxicity (LD50). 
Unless otherwise noted, we report AUROC for classification and RMSE (↓) for regression, following common practice; dataset details and splits are in Appendix~\ref{appendix:task_description}. Also, in order to easy to reproduce, we give the hyper-parameters we used in Appendix~\ref{appendix:exp_setup_paras}.

\subsection{Baseline}
To comprehensively evaluate our ChemATP framework, we consider a broad set of baseline methods spanning training-free, LLM-based, graph-based, and 3D molecular modeling approaches. Specifically, we include training-free methods such as MolRAG~\citep{xian2025molrag} and LLM4SD~\citep{llm4sd}. For LLMs with task-specific training, we compare against InstructMol~\citep{cao2023instructmol} and MoMu~\citep{momus}. In addition, we incorporate representative graph-based baselines, including GraphMVP~\citep{liu2022pretraining} and MoleculeSTM~\citep{liu2023multi}, as well as the 3D molecular modeling methods UniMol~\citep{zhou2023uni} and 
TokenMol~\citep{wang2025token}.

\input{results/ablation}

\subsection{Main Results}
\label{sec:main_results}
We summarize the overall performance against both training-free and training-based baselines in Table~\ref{tab:main_results}. 
These findings provide a definitive affirmative answer to \textbf{RQ1}. 
First, ChemATP consistently achieves the best performance among training-free methods, significantly outperforming baselines that rely on surface-level prompting. 
Second, and more notably, it successfully rivals and in several cases surpasses representative training-based approaches such as MoMu~\citep{momus}, InstructMol~\citep{cao2023instructmol}. 
This demonstrates that explicitly injecting fine-grained priors is not merely a compromise for efficiency, but a robust strategy that allows frozen LLMs to match the predictive power of fully fine-tuned graph encoders.

Beyond numerical accuracy, ChemATP offers a qualitative advantage in transparency (example outputs in Appendix~\ref{appendix:classification_task_output}). 
Unlike end-to-end neural baselines where decision boundaries are opaque, ChemATP provides traceable reasoning paths. 
By inspecting the model's articulated reasoning regarding the selected priors and their stated implications for the prediction, domain practitioners can verify whether the decision is based on sound chemical principles or spurious correlations.
This transparency fosters trust, ensuring that high performance is achieved for the right reasons. 

We further evaluate ChemATP on regression tasks; complete results are provided in Appendix~\ref{appendix:additional_results}.

\subsection{Ablation Study}
We conduct ablation studies to isolate the impact of two core design components: (i) the contribution of multi-level textualized priors, and (ii) the role of retrieved exemplars in grounding reasoning.

\subsubsection{Contribution of Multi-Level Priors}
\label{sec:abla_on_atomic_priors}
To answer \textbf{RQ2}, we compare the full ChemATP against variants that strip away specific prior levels while keeping other settings fixed. \emph{w/o Atom. Pr.} removes the atom-level textualized priors, and \emph{w/o Mol. Pr.} removes the molecule-level descriptors.

Results in Table~\ref{tab:ablation_chematp} reveal that the full ChemATP consistently achieves the best performance, confirming that atom- and molecule-level priors are not redundant but synergistic.
Specifically, removing atom-level priors causes marked degradation on datasets driven by specific pharmacophores or local environments (e.g., lower AUROC on BACE, higher RMSE on FreeSolv), highlighting the necessity of fine-grained local information.
Conversely, removing molecule-level priors also hurts performance (e.g., on BBBP and ClinTox), suggesting that global physicochemical context (e.g., solubility, weight) provides essential macroscopic constraints.
The aggregate trend demonstrates that integrating these two levels of granularity yields the most robust reasoning, validating our design of a comprehensive, multi-level knowledge base.

\subsubsection{Impact of Retrieval and Zero-Shot Robustness}
\label{sec:abla_on_top_k}
\input{results/zero_shot}
\input{results/diff_LLMs}
To address \textbf{RQ3}, we examine the effect of analogue retrieval by varying $k \in \{0,2,4,6,8,10\}$. Here, $k{=}0$ represents the Zero-Shot setting, where ChemATP relies solely on its internal knowledge base without external exemplars.

Table~\ref{tab:ablation_chematp} shows a clear trend: moving from $k{=}0$ to a small positive $k$ yields substantial gains. This confirms that retrieved analogues successfully activate the LLM's capability for reasoning by analogy, allowing it to validate hypotheses against concrete examples. Notably, performance saturates around $k{\approx}4$--$6$, suggesting that ChemATP is highly efficient—it does not require a large crowd of examples, but rather a handful of high-quality anchors to ground its decision.

We further evaluate whether ChemATP remains effective without any exemplars. As shown in Table~\ref{tab:zero-shot}, Zero-Shot ChemATP (ZS-ChemATP) is competitive with, and often superior to, standard few-shot baselines. 
Remarkably, ZS-ChemATP outperforms Few-Shot CoT on nearly all datasets and surpasses Few-Shot MolRAG on \textbf{4/5} comparable benchmarks.
This result is particularly significant: it implies that access to a high-quality explicit knowledge base allows the model to reason effectively even without specific examples, validating our core premise that explicit grounding can effectively substitute or outperform for surface-level pattern matching from few-shot prompting.

\subsection{Generalization across LLM Backbones}
\label{sec:diff_llms}
To verify the universality of our framework, we instantiate ChemATP across a diverse spectrum of LLM backbones—ranging from efficient distilled models (GPT-5-Nano) to frontier-class reasoning engines (DeepSeek-V3.2, GPT-5, GPT-4.1, Claude Sonnet-4). All experiments use identical prompts and retrieval pipelines without specific tuning.

Table~\ref{tab:diff_llms} reveals a universal enhancement pattern. 
ChemATP consistently achieves the best average performance across all backbones compared to standard prompting strategies (DA, CoT, and their few-shot variants) for both classification and regression tasks. 
Crucially, we observe a democratizing effect: smaller models equipped with ChemATP frequently rival or even surpass larger models relying on standard prompting. 
This suggests that the bottleneck in chemical reasoning is often the lack of explicit domain priors rather than raw parametric capacity.
These results affirmatively answer \textbf{RQ4}: ChemATP is a robust, backbone-agnostic framework. 
By decoupling knowledge from the model, it serves as a plug-and-play enhancement layer that upgrades any LLM—regardless of vendor, architecture, or scale—into a grounded chemical reasoner.

\begin{figure*}[!t]
  \centering
  \includegraphics[width=\textwidth]{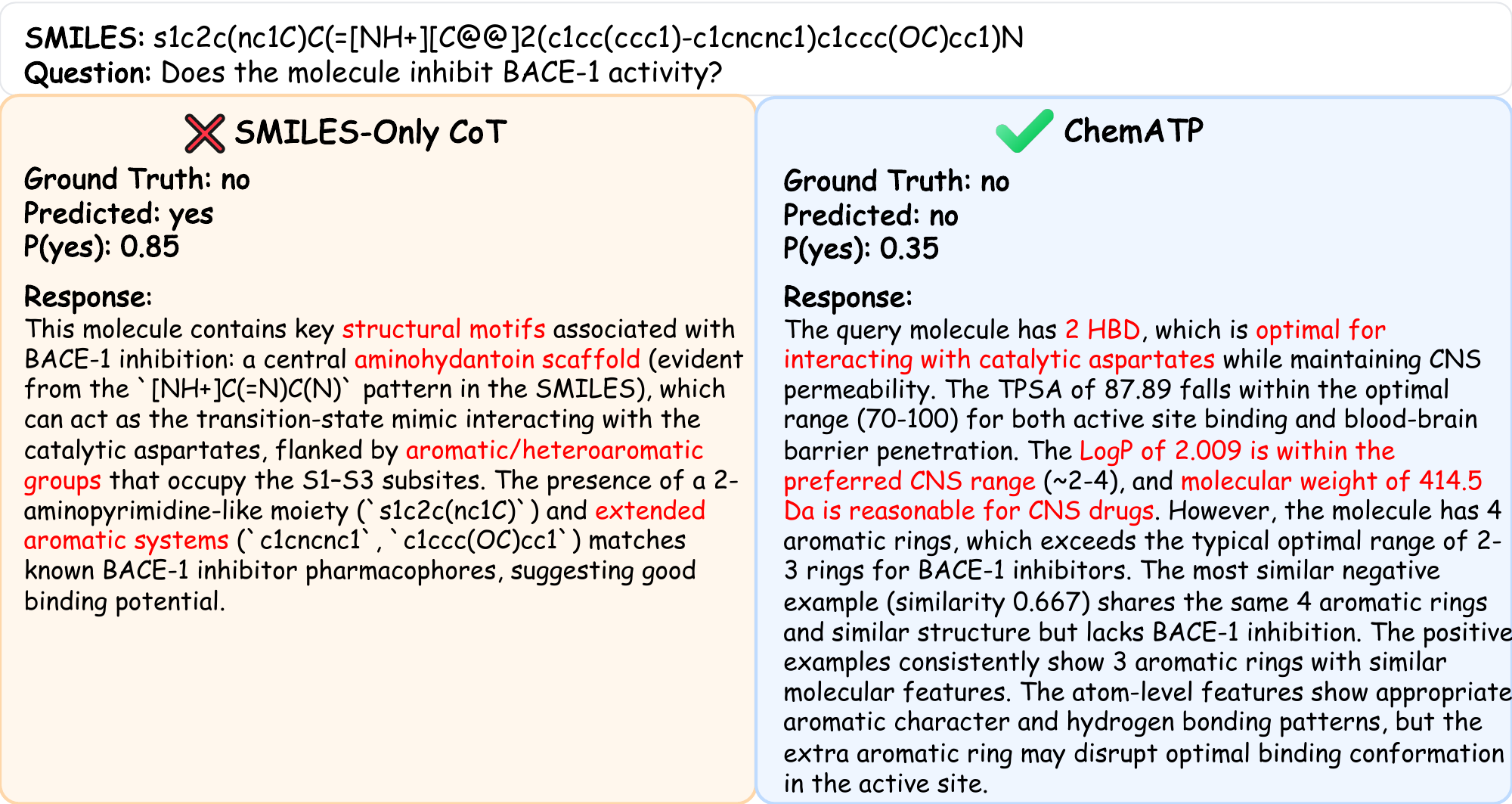}
\caption{Comparison between SMILES-only CoT and ChemATP on a BACE-1 inhibition task. 
The left panel illustrates how the baseline falls into a hallucination trap, deriving an overconfident prediction from surface-level structural motifs without physical verification. 
In contrast, the right panel demonstrates ChemATP's evidence-based reasoning: by explicitly citing quantitative priors (e.g., TPSA, LogP), it successfully rejects the candidate based on physicochemical constraints. 
Red highlights denote the verifiable attributes guiding the correct decision.}
\label{fig:chematp_cot_comparison}
\end{figure*}

\subsection{Case Study: From Surface Heuristics to Evidence-Based Verification}
\label{sec:case_study}
To demonstrate how ChemATP utilizes explicit priors to ground the reasoning process, Fig.~\ref{fig:chematp_cot_comparison} presents a side-by-side comparison on a BACE-1 inhibition task. This comparison illustrates the shift from implicit pattern matching to quantitative verification.

The SMILES-only baseline (left) fails due to the opacity of the 1D string representation. 
While it correctly identifies a potentially active scaffold, it lacks the granular information necessary to verify the molecule's physicochemical compatibility. 
Constrained by this information deficit, the model relies on heuristic-driven inference, falsely associating the mere presence of a drug-like scaffold with biological activity. 

In contrast, ChemATP (right) equips the LLM with specific, textualized priors such as the precise aromatic ring count, TPSA, and LogP which serve as quantitative evidence for reasoning. 
This explicit grounding enables the model to validate the molecule against established structure-activity relationships. 
By shifting the paradigm from logical hallucination to fact-conditioned reasoning, ChemATP successfully rejects the superficially promising candidate based on rigorous physicochemical constraints.

\section{Conclusion}
In this work, we propose ChemATP, a framework that redefines molecular intelligence by \textbf{decoupling} chemical knowledge from the reasoning engine. 
Instead of relying on the static, opaque parameters of training-based models, ChemATP externalizes knowledge into the first atom-level textual knowledge base. 
This architecture empowers LLMs to perform grounded reasoning by explicitly retrieving fine-grained atomic evidence and molecule-level descriptors, ensuring both adaptability and interpretability. 
Empirically, we demonstrate that this paradigm is not merely a resource-efficient alternative but a competitive or even a superior strategy: ChemATP significantly outperforms training-free baselines and rivals state-of-the-art training-based methods across diverse benchmarks. 
By proving that explicit prior injection can effectively substitute for implicit parameter updates, ChemATP establishes a new path for transparent and robust AI in scientific discovery.

\section{Limitations}
Although we achieve strong performance on molecular property prediction tasks and provide better explanations, several limitations remain. First, we conduct experiments only on property-prediction benchmarks; in future work, we plan to apply ChemATP to more complex tasks, such as retrosynthesis. Second, the current pipeline feeds many tokens into the model, and more efficient filtering methods could reduce the token budget. Lastly, we plan to extract potential scientific insights from the responses generated by the our framework.

\newpage
\bibliography{custom}
\clearpage
\newpage
\appendix

\section{Experimental Details}
\subsection{Task Descriptions}
\label{appendix:task_description}

\paragraph{BBBP.} This task assesses whether a compound can cross the blood-brain barrier (BBB), a crucial consideration for central nervous system (CNS) drugs. Molecules are classified into BBB-permeable or non-permeable categories based on experimental permeability data.

\paragraph{Tox21.} A multi-label classification task involving 12 distinct toxicity pathways. These include nuclear receptor signaling and stress response pathways, reflecting the compound’s potential toxic effects across multiple biological processes.

\paragraph{ToxCast.} This task covers a broader panel of in vitro toxicity assays than Tox21, measuring compound bioactivity across hundreds of biochemical and cellular targets. It enables high-throughput evaluation of toxicity at scale.

\paragraph{SIDER.} The Side Effect Resource dataset includes marketed drugs annotated with known adverse drug reactions. The goal is to predict side effect profiles across 27 categories, making this a multi-label classification task.

\paragraph{ClinTox.} This binary classification task distinguishes drugs approved by the FDA from those that failed clinical trials due to toxicity. It provides a stringent benchmark for modeling toxicity-related risk.

\paragraph{HIV.} The HIV dataset involves binary classification of compounds based on their ability to inhibit HIV replication in a human cell assay. It evaluates antiviral activity and molecular efficacy.

\paragraph{BACE.} This task aims to predict inhibitors of human $\beta$-secretase 1 (BACE-1), an enzyme linked to Alzheimer’s disease. Compounds are labeled based on their biochemical assay outcomes.

\paragraph{QM9.} From the QM9 quantum chemistry dataset, we focus on predicting three scalar properties: the energies of the Highest Occupied Molecular Orbital (HOMO), the Lowest Unoccupied Molecular Orbital (LUMO), and their difference (the HOMO–LUMO gap). These properties are central to understanding molecular reactivity, stability, and electronic behavior.

\paragraph{ESOL.} This task predicts the aqueous solubility of small molecules. ESOL provides experimental log solubility values for a curated set of drug-like compounds.

\paragraph{FreeSolv.} FreeSolv contains hydration free energy values for small neutral molecules. The task evaluates the model’s ability to predict solvation thermodynamics, crucial for drug absorption modeling.

\paragraph{Lipo.} The Lipophilicity dataset measures the distribution coefficient ($\log D$) between octanol and water. It reflects how hydrophobic or hydrophilic a compound is, which is relevant to absorption and bioavailability.

\paragraph{Caco2.} This task involves predicting the permeability of molecules across Caco-2 cell monolayers, a common in vitro assay used to estimate human intestinal absorption.

\paragraph{LD50.} LD50 measures the lethal dose required to kill 50\% of test subjects (typically rodents), serving as a proxy for acute toxicity prediction.

\paragraph{AqSol.} AqSol contains experimentally measured solubility values for a broad set of compounds, aiming to improve solubility prediction beyond the chemical space covered by ESOL.

\subsection{Setup for Filtering Atom Cards}
\label{appendix:atom_card_selection}

We do \emph{not} pass all atom cards to Stage~III. For each molecule we keep only a small, decision-critical subset. Concretely, we first mark \textbf{must-keep} atoms: any RDKit-identified hydrogen-bond donor/acceptor (from the ChemicalFeatures \texttt{BaseFeatures.fdef}), any atom with nonzero formal charge or \(|\)Gasteiger\(|\!\ge\!0.10\), all heteroatoms (N, O, S, P, F, Cl, Br, I), and ring bridgeheads/aromatic carbons. If must-keep atoms exceed a per-molecule budget \(M\) (default \(M\!=\!20\)), we retain those with larger \(|\)Gasteiger\(|\) and aromaticity first (stable ties by atom index). If they are fewer than \(M\), we fill the remaining slots by ranking the rest with a light priority that favors aromatic/conjugated sites, ring/fused-ring membership, proximity to heteroatoms, and \texttt{sp}/\texttt{sp2} hybridization. The final selection is serialized as compact cards (token id, atom index, symbol, plus only the Stage~I–selected fields), which focuses the prompt on functional centers instead of flooding the context with non-informative atoms.

\subsection{Experimental Setup and Parameters}
\label{appendix:exp_setup_paras}

For all datasets, we follow the standard MoleculeNet evaluation protocol by holding out 10\% of the data as the test set and \emph{not} training any task-specific model; the training split is used exclusively as the retrieval pool in StageII. Following MoleculeNet\citep{wu2018moleculenet}, we use scaffold splits for the recommended tasks and random splits otherwise, with a fixed random seed of 42. Molecular similarity is computed using RDKit Morgan fingerprints and Tanimoto similarity, and we retrieve Top-$k$ analogues (including both positive and negative examples) for each test molecule. We query LLMs with deterministic decoding (temperature~=0) and a fixed maximum output length; all other decoding parameters are left at provider defaults. We evaluate classification tasks using AUROC and regression tasks using RMSE. Because logits are inaccessible for closed-source LLMs, we use the model-reported probability $P(\text{yes}) \in [0,1]$ as the prediction score when computing AUROC. Full settings and hyperparameters are provided in Table~\ref{tab:exp_hparams}.

\begin{table}[htbp]
\centering
\scriptsize
\setlength{\tabcolsep}{4pt}
\begin{tabular}{lll}
\toprule
\textbf{Component} & \textbf{Hyperparameter} & \textbf{Value / Setting} \\
\midrule
Data split & Train:test ratio & 90:10 \\
          & Random seed       & 42  \\
\midrule
Retrieval (Stage II) & Fingerprint type & RDKit Morgan (ECFP4) \\
                     & Radius           & 2 \\
                     & \texttt{nBits}   & 2048 \\
                     & Similarity       & Tanimoto \\
                     & Top-$k$ analogues & $k = 5$ or $k = 10$ \\
\midrule
LLM decoding & Temperature      & 0.0 \\
             & Max tokens       & 2000 \\
             & Top-$p$          & Default \\
             & Other decoding   & Default \\
\midrule
Metrics & Classification        & AUROC\\
        & Regression            & RMSE \\
\bottomrule
\end{tabular}
\caption{Summary of experimental hyperparameters and implementation choices used in ChemATP.}
\label{tab:exp_hparams}
\end{table}

\subsection{Baselines}

To comprehensively evaluate our ChemATP framework, we compare against both \emph{training-free} and \emph{training-based} baselines. For training-based methods, we further include \emph{LLM-based}, \emph{graph-based}, and \emph{3D-based} approaches. All baseline results are reported as in their original papers.

\paragraph{Training-free methods.}
We include MolRAG~\citep{xian2025molrag} and LLM4SD~\citep{llm4sd}. 
MolRAG computes the similarity between a test molecule and the training set, then supplies the similarity scores and the corresponding training labels to LLMs as in-context evidence, allowing the model to make decisions by analogy. 
LLM4SD prompts LLMs to induce classification rules; RDKit-derived, molecule-level chemical priors are converted into vector features, upon which a decision tree is trained for prediction.

\paragraph{LLM-based training methods.}
We consider InstructMol~\citep{cao2023instructmol}, TokenMol~\citep{wang2025token}, and MoMu~\citep{momus}. 
InstructMol trains an additional adapter that projects molecular representations into the LLM embedding space; the projected vector is concatenated with the original prompt and injected into the transformer, so the LLM learns to leverage these features for decision making. 
MoMu performs contrastive learning between text embeddings and molecular embeddings to learn a unified cross-modal representation. 
TokenMol discretizes 3D features—specifically torsion angles—into tokens, concatenates them with the prompt to form a token-only sequence, and trains the model via random-mask autoregression.

\paragraph{Graph-based methods.}
We adopt GraphMVP~\citep{liu2022pretraining}, which conducts contrastive learning between 2D (GIN-based) and 3D (SchNet-based~\citep{schutt2017schnet}) molecular representations to capture complementary structural information.

\paragraph{3D-based methods.}
We include UniMol~\citep{zhou2023uni}, which pretrains on molecular conformations and protein pockets, leveraging rotation–translation-invariant spatial positional encodings and achieving state-of-the-art performance on molecular property prediction.

\subsection{Knowledge Base Construction Details}

We construct the atomic-prior knowledge base from a subset of approximately 220{,}000 molecules sampled from PubChem, using only their SMILES strings. Specifically, we encode each molecule with froze MoleculeSTM checkpoint to obtain atom-level representations, and then map these representations to discrete token indices via AtomDisc. Importantly, this entire pipeline is performed in a fully unsupervised manner: we do not use any labels or annotations from downstream property-prediction tasks, nor do we tune any component on evaluation datasets. So we don't need to train any models in our framework. For all tasks, we only use out-off-shelf models' checkpoint to obtain the necessary information and textualize them using rkdit and our methods.

To further avoid any potential data leakage, the knowledge base is built solely from the PubChem subset and is kept completely disjoint from all MoleculeNet evaluation datasets; no molecules from the test/validation splits are used during knowledge base construction. Atom-level priors are computed purely from molecular structures using RDKit-based descriptors and statistics, without referencing any task targets.

For reproducibility and community use, we publicly release the complete ChemATP framework, including the knowledge-base construction code and prompting scripts, at \texttt{https://github.com/mingxuZhang2/ChemATP}.

\subsection{Evaluation Details}

This section describes our evaluation protocol for classification and regression tasks.

\paragraph{Classification.}
For closed-source LLMs, token-level logits are not accessible. Therefore, instead of computing AUROC from logits, we prompt the model to output a calibrated confidence score for the positive class, i.e., $p(\text{yes}) \in [0,1]$, in a structured format. Concretely, the model returns a numeric value wrapped by \texttt{<confidence>} and \texttt{</confidence>}, and we parse this value as the prediction score to compute AUROC.
For multi-label (or multi-class) settings, we decompose each task into a set of binary one-vs-rest (or per-label) sub-tasks, and report AUROC for each label/item accordingly. 

\paragraph{Regression.}
For regression tasks, we compute RMSE directly between the model predictions and the dataset-provided ground-truth labels. We do not apply any additional target transformations (e.g., log-space conversion) or target standardization/normalization. Since our approach is training-free, we evaluate in the original label space as provided by each benchmark dataset, and compute RMSE on this same scale.
To avoid ambiguity, we will also explicitly state (per dataset) the target definition and units (when available) in the dataset description or appendix.

\clearpage
\newpage

\section{Atomic Prior Knowledge Base Schema}
\label{appendix:kb_schema}
\input{prompt/schema_example}
\clearpage
\newpage
\section{Input Format Example for Retrieved Analogues}
\label{appendix:example_input}

\input{prompt/input_analogy_example}

\section{Input Format for Tested Molecules}
\label{appendix:example_input_tested_sample}

\input{prompt/input_tested_molecule}

\clearpage
\newpage
\section{Prompt Templates}
\label{sec:prompts}

\subsection{Prompt Template for Stage I}
\label{appendix:prompt_example_stage1}

\input{prompt/stage1}
\clearpage
\newpage

\subsection{Prompt Template for Stage III}
\label{appendix:prompt_example_stage3}

\input{prompt/stage3}
\clearpage
\newpage

\subsection{Baseline Prompt}
\label{appendix:baseline_prompt}
\input{prompt/DA}

\input{prompt/CoT}

\clearpage
\newpage
\section{Output Example}

\subsection{Stage I Output Example}
\subsubsection{Stage I-Round 1 Example}
\input{output/stage1-round1}

\clearpage
\newpage
\subsection{Classification Task Output Example}
\label{appendix:classification_task_output}
\input{output/example_output_classification_bace}
\input{output/example_output_classification_bbbp}
\input{output/example_output_classification_clintox}
\clearpage
\newpage
\input{output/example_output_classification_hiv}

\input{output/example_output_classification_tox21}
\input{output/example_output_classification_sider}
\clearpage
\newpage

\onecolumn
\section{Additional Experimental Results}
\label{appendix:additional_results}
\input{results/regression}
\begin{table*}[htbp]
\centering
\scriptsize
\setlength{\tabcolsep}{6pt}
\renewcommand{\arraystretch}{1.12}
\caption{Regression benchmarks (RMSE $\downarrow$). Each dataset is shown as a multirow block with per-LLM results. The bottom row reports the average across all regression datasets and LLMs.}
\label{tab:appendix_regression_multirow}
\resizebox{\textwidth}{!}{%
\begin{tabular}{l l ccccc}
\toprule
\textbf{Dataset} & \textbf{Model} & \textbf{ChemATP} & \textbf{DA+FS} & \textbf{CoT+FS} & \textbf{DA} & \textbf{CoT} \\
\midrule
\multirow{5}{*}{\textbf{ESOL}} 
& \texttt{Claude-Sonnet-4} & 0.5854 & 0.9414 & 0.9043 & 0.8589 & 0.9654 \\
& \texttt{DeepSeek-V3.2}   & 0.6504 & 1.0497 & 0.9134 & 1.0194 & 0.8944 \\
& \texttt{GPT-4.1}         & 0.6940 & 1.1039 & 0.9846 & 1.0487 & 0.9363 \\
& \texttt{GPT-5}           & 0.5900 & 0.8114 & 0.8529 & 0.8833 & 0.8668 \\
& \texttt{GPT-5-Nano}      & 0.7729 & 1.4503 & 1.4259 & 1.3622 & 1.3085 \\
\midrule
\multirow{5}{*}{\textbf{FreeSolv}} 
& \texttt{Claude-Sonnet-4} & 0.9366 & 2.5170 & 2.9579 & 3.3703 & 3.3429 \\
& \texttt{DeepSeek-V3.2}   & 1.0177 & 2.5510 & 3.2863 & 2.4534 & 3.4312 \\
& \texttt{GPT-4.1}         & 1.0364 & 5.2186 & 3.9521 & 8.9247 & 11.4176 \\
& \texttt{GPT-5}           & 0.9930 & 2.6808 & 3.1213 & 4.2696 & 4.0424 \\
& \texttt{GPT-5-Nano}      & 1.0009 & 3.0818 & 2.9089 & 5.8587 & 6.8408 \\
\midrule
\multirow{5}{*}{\textbf{Lipophilicity}} 
& \texttt{Claude-Sonnet-4} & 0.7107 & 1.2095 & 1.1821 & 1.2493 & 1.3552 \\
& \texttt{DeepSeek-V3.2}   & 0.7143 & 1.4555 & 1.2479 & 1.3517 & 1.2098 \\
& \texttt{GPT-4.1}         & 0.7338 & 1.2383 & 1.2615 & 1.2134 & 1.2376 \\
& \texttt{GPT-5}           & 0.6227 & 1.0466 & 1.0114 & 1.0542 & 1.0792 \\
& \texttt{GPT-5-Nano}      & 0.7051 & 1.4405 & 1.3494 & 1.3876 & 1.3302 \\
\midrule
\textbf{Avg. RMSE~$\downarrow$} & & \textbf{0.7842} & 1.8531 & 1.8240 & 2.4204 & 2.6839 \\
\bottomrule
\end{tabular}%
}
\end{table*}

\input{results/knn}

\begin{table*}[htbp]
\centering
\scriptsize
\setlength{\tabcolsep}{6pt}
\renewcommand{\arraystretch}{1.12}
\caption{Classification benchmarks (AUROC $\uparrow$). Each dataset is shown as a multirow block with per-LLM results. The bottom row reports the average across all classification datasets and LLMs.}
\label{tab:appendix_classification_multirow}
\resizebox{\textwidth}{!}{%
\begin{tabular}{l l ccccc}
\toprule
\textbf{Dataset} & \textbf{Model} & \textbf{ChemATP} & \textbf{DA+FS} & \textbf{CoT+FS} & \textbf{DA} & \textbf{CoT} \\
\midrule
\multirow{5}{*}{\textbf{BACE}} 
& \texttt{Claude-Sonnet-4} & 0.8028 & 0.6218 & 0.6694 & 0.6891 & 0.6390 \\
& \texttt{DeepSeek-V3.2}   & 0.8042 & 0.6494 & 0.6373 & 0.6428 & 0.6775 \\
& \texttt{GPT-4.1}         & 0.8443 & 0.7087 & 0.7608 & 0.7330 & 0.7012 \\
& \texttt{GPT-5}           & 0.8970 & 0.6941 & 0.7106 & 0.7484 & 0.6975 \\
& \texttt{GPT-5-Nano}      & 0.6729 & 0.5709 & 0.6773 & 0.5984 & 0.6421 \\
\midrule
\multirow{5}{*}{\textbf{BBBP}} 
& \texttt{Claude-Sonnet-4} & 0.8173 & 0.7604 & 0.7730 & 0.7315 & 0.7686 \\
& \texttt{DeepSeek-V3.2}   & 0.8246 & 0.5510 & 0.6660 & 0.4918 & 0.6121 \\
& \texttt{GPT-4.1}         & 0.8074 & 0.7335 & 0.7584 & 0.7304 & 0.7570 \\
& \texttt{GPT-5}           & 0.8731 & 0.8856 & 0.8340 & 0.8685 & 0.8435 \\
& \texttt{GPT-5-Nano}      & 0.7259 & 0.4175 & 0.4439 & 0.3435 & 0.4081 \\
\midrule
\multirow{5}{*}{\textbf{ClinTox}} 
& \texttt{Claude-Sonnet-4} & 0.7614 & 0.4030 & 0.5085 & 0.4870 & 0.5353 \\
& \texttt{DeepSeek-V3.2}   & 0.8163 & 0.7792 & 0.6283 & 0.8459 & 0.8902 \\
& \texttt{GPT-4.1}         & 0.7514 & 0.5584 & 0.5589 & 0.3223 & 0.2887 \\
& \texttt{GPT-5}           & 0.8856 & 0.5799 & 0.6033 & 0.7524 & 0.5258 \\
& \texttt{GPT-5-Nano}      & 0.6927 & 0.4301 & 0.5075 & 0.1574 & 0.4769 \\
\midrule
\textbf{Avg. AUROC~$\uparrow$} & & \textbf{0.7985} & 0.6229 & 0.6492 & 0.6095 & 0.6309 \\
\bottomrule
\end{tabular}%
}
\end{table*}

\end{document}

%% file: results/classification.tex
\begin{table*}[!t]
\centering
\scriptsize
\setlength{\tabcolsep}{3pt}
\renewcommand{\arraystretch}{1.15}
\caption{Comparison of ChemATP with baselines on classification benchmarks (AUROC). \emph{DA} prompts LLMs to answer directly from the given molecule, whereas \emph{CoT} prompts LLMs to explicitly reason about the molecule. Prompt templates for both settings are provided in Appendix~\ref{appendix:baseline_prompt}. All baseline results are taken as reported in their respective papers and entries marked ``---'' were not reported in the original papers.}
\label{tab:main_results}
\resizebox{\textwidth}{!}{%
\begin{tabular}{lccccccccc}
\toprule
\textbf{Dataset}
& \textbf{ChemATP}
& \textbf{DA}
& \textbf{CoT}
& \textbf{MolRAG}
& \textbf{LLM4SD}
& \textbf{InstructMol}
& \textbf{MoMu}
& \textbf{GraphMVP}
& \textbf{UniMol} \\
\midrule
BACE    & \textbf{0.8970} & 0.6428 & 0.6775 & 0.7673 & 0.8383 & 0.8230 & 0.7670 & 0.8120 & 0.8570 \\
BBBP    & \textbf{0.8731} & 0.4918 & 0.6121 & 0.6291 & 0.7560 & 0.7000 & 0.7050 & 0.7240 & 0.7290 \\
ClinTox & {0.8856} & 0.5015 & 0.4860 & ---    & 0.9130 & 0.9150 & 0.7990 & 0.7910 & \textbf{0.9190} \\
HIV     & {0.7270} & 0.6452 & 0.6266 & 0.6476 & 0.7901 & 0.6890 & 0.7590 & 0.7700 & \textbf{0.8080} \\
SIDER   & \textbf{0.7179} & 0.6371 & 0.6279 & ---    & 0.6540 & 0.5780 & 0.6050 & 0.6390 & 0.6590 \\
Tox21   & \textbf{0.8023} & 0.5685 & 0.5958 & 0.6642 & 0.7470 & 0.7470 & 0.7560 & 0.7590 & 0.7960 \\
\midrule
\textbf{Avg}
& \textbf{0.8172}
& 0.5812
& 0.6043
& 0.6771
& 0.7831
& 0.7420
& 0.7318
& 0.7492
& 0.7947 \\
\bottomrule
\end{tabular}%
}
\end{table*}

%% file: results/ablation.tex
\begin{table*}[!t]
\centering
\scriptsize
\setlength{\tabcolsep}{4pt}
\renewcommand{\arraystretch}{1.15}
\caption{Ablation study of ChemATP along two axes: (i) the contribution of chemical priors (atom and molecule level) and (ii) the effect of the Top-$k$ selection. Classification results (BACE, BBBP, ClinTox) are reported as AUROC~($\uparrow$), and regression results (Caco-2, FreeSolv, Lipophilicity) as RMSE~($\downarrow$).}
\label{tab:ablation_chematp}
\resizebox{\textwidth}{!}{%
\begin{tabular}{lccccccccc}
\toprule
\textbf{Dataset} & \textbf{ChemATP} & \textbf{w/o Atom. Pr.} & \textbf{w/o Mol. Pr.} & \textbf{$k=0$} & \textbf{$k=2$} & \textbf{$k=4$} & \textbf{$k=6$} & \textbf{$k=8$} & \textbf{$k=10$} \\
\midrule
\textbf{BACE}      & 0.8042 & 0.7290 & 0.8207 & 0.6694 & 0.6594 & 0.8223 & 0.8222 & \textbf{0.8511} & 0.8042 \\
\textbf{BBBP}      & \textbf{0.8246} & 0.8131 & 0.7176 & 0.7887 & 0.7712 & 0.8095 & 0.8094 & 0.8181 & {0.8246} \\
\textbf{ClinTox}   & \textbf{0.8163} & 0.7974 & 0.7072 & 0.5902 & 0.7288 & 0.8018 & 0.7741 & 0.7664 & {0.8163} \\
\midrule
\textbf{Avg. AUROC} & \textbf{0.8150} & 0.7798 & 0.7485 & 0.6828 & 0.7198 & 0.8112 & 0.8019 & 0.8119 & {0.8150} \\
\midrule
\textbf{Caco-2}    & {0.5018} & 0.5271 & 0.5028 & 2.2123 & 0.5825 & 0.5261 & 0.5018 & \textbf{0.4935} & 0.5041 \\
\textbf{FreeSolv}  & \textbf{0.9008} & 1.0210 & 0.9776 & 2.8607 & 1.2379 & 1.0253 & 0.9008 & 1.0540 & 1.0177 \\
\textbf{Lipo}      & \textbf{0.7022} & 0.7917 & 0.7844 & 1.1916 & 0.7679 & 0.7674 & 0.7022 & 0.7327 & 0.7143 \\
\midrule
\textbf{Avg. RMSE}  & \textbf{0.7016} & 0.7799 & 0.7549 & 2.0882 & 0.8628 & 0.7729 & 0.7016 & 0.7601 & 0.7454 \\
\bottomrule
\end{tabular}%
}
\end{table*}

%% file: results/zero_shot.tex

\begin{table}[htbp]
\centering
\scriptsize
\setlength{\tabcolsep}{4pt}
\renewcommand{\arraystretch}{1.12}
\caption{Comparison across zero-shot ChemATP (ZS-ChemATP), few-shot MolRAG (FS-MolRAG), and chain-of-thought prompting in zero-shot (ZS-CoT) and few-shot (FS-CoT) settings. Classification performance is reported as AUROC~($\uparrow$), and regression performance as RMSE~($\downarrow$). The best results are shown in \textbf{bold}, and the runner-up results are \underline{underlined}.}
\label{tab:zero-shot}
\begin{tabular}{lcccc}
\toprule
\textbf{Dataset} & \textbf{ZS-ChemATP} & \textbf{FS-MolRAG} & \textbf{ZS-CoT} & \textbf{FS-CoT} \\
\midrule
\multicolumn{5}{l}{\emph{Classification (AUROC $\uparrow$)}} \\
BACE         & 0.6694  & \textbf{0.7673} & \underline{}{0.6475} & 0.6373 \\
BBBP         & \textbf{0.7887}  & 0.6291 & 0.6121 & \underline{}{0.6660} \\
ClinTox      & \underline{0.5902}  & ---    & 0.4860 & \textbf{0.6283} \\
\midrule
\multicolumn{5}{l}{\emph{Regression (RMSE $\downarrow$)}} \\
ESOL         & \textbf{0.6315}  & 3.2806 & 0.8944 & 0.9134 \\
FreeSolv     & \textbf{2.8607}  & 6.0827 & 3.4312 & 3.2863 \\
Lipophilicity& \textbf{1.1013}  & 1.1251 & 1.2098 & 1.2479 \\
\bottomrule
\end{tabular}
\end{table}

%% file: results/diff_LLMs.tex
\begin{table*}[t]
\centering
\scriptsize
\setlength{\tabcolsep}{6pt}
\renewcommand{\arraystretch}{1.12}
\caption{Results across different LLMs integrated with ChemATP on classification (AUROC$\uparrow$) and regression (RMSE$\downarrow$) tasks. To highlight the advantage of explicit chemical priors over single-SMILES prompting, we compare four inference strategies: DA+FS, CoT+FS, DA, and CoT. Here, \emph{DA} denotes directly answering from the prompt, \emph{CoT} denotes reasoning, and \emph{FS} indicates few-shot examples.}
\label{tab:diff_llms}
\resizebox{\textwidth}{!}{%
\begin{tabular}{l l ccccc}
\toprule
\textbf{Task (Avg over datasets)} & \textbf{LLM} 
& \textbf{ChemATP} & \textbf{DA+FS} & \textbf{CoT+FS} & \textbf{DA} & \textbf{CoT} \\
\midrule
\multirow{6}{*}{\textbf{Regression}}
& \texttt{Claude-Sonnet-4} & \textbf{0.7442} & 1.5560 & 1.6815 & 1.8262 & 1.8878 \\
& \texttt{Deepseek-V3.2}             & \textbf{0.7941} & 1.6854 & 1.8159 & 1.6081 & 1.8451 \\
& \texttt{GPT-4.1}                   & \textbf{0.8214} & 2.5202 & 2.0661 & 3.7289 & 4.5305 \\
& \texttt{GPT-5}          & \textbf{0.7352} & 1.5129 & 1.6619 & 2.0690 & 1.9961 \\
& \texttt{GPT-5-Nano}     & \textbf{0.8263} & 1.9909 & 1.8948 & 2.8695 & 3.1598 \\
\cmidrule(lr){2-7}
& \textbf{Avg. RMSE~$\downarrow$}           & \textbf{0.7842} & 1.8531 & 1.8240 & 2.4204 & 2.6839 \\
\midrule
\multirow{6}{*}{\textbf{Classification}}
& \texttt{Claude-Sonnet-4} & \textbf{0.7938} & 0.5950 & 0.6503 & 0.6359 & 0.6476 \\
& \texttt{Deepseek-V3.2}            & \textbf{0.8150} & 0.6599 & 0.6439 & 0.6601 & 0.7266 \\
& \texttt{GPT-4.1}                  & \textbf{0.8010} & 0.6669 & 0.6927 & 0.5952 & 0.5823 \\
& \texttt{GPT-5}         & \textbf{0.8852} & 0.7199 & 0.7159 & 0.7898 & 0.6889 \\
& \texttt{GPT-5-Nano}    & \textbf{0.6972} & 0.4728 & 0.5429 & 0.3664 & 0.5090 \\
\cmidrule(lr){2-7}
& \textbf{Avg. AUROC~$\uparrow$}          & \textbf{0.7985} & 0.6229 & 0.6492 & 0.6095 & 0.6309 \\
\bottomrule
\end{tabular}%
}
\end{table*}

%% file: prompt/schema_example.tex
\begin{chemprompt}{Atomic Prior KB Schema — Fields \& Example}
Schema:
- token_id: discrete AtomDisc token index
- support_count: #atoms in corpus mapped to this token
- primary_symbol: most frequent element symbol among instances
- is_mixed: whether multiple element types map to this token
- symbol_distribution: per-element counts for this token
- mixture_entropy: entropy (0–1) of the element mixture
- env_type: modal local environment {chain|ring|fused_ring}
- env_distribution: counts of {chain, ring, fused_ring}
- aromatic_ratio: fraction of instances that are aromatic (0–1)
- conjugated_ratio: fraction with any conjugated bond (0–1)
- median_degree: median number of bonded neighbors
- median_ring_size: median size of the smallest ring (0 if non-ring)
- hybridization: modal hybridization {s, sp, sp2, sp3, sp3d, sp3d2, other}
- electrics.inductive: median inductive effect sign in {-1,0,+1}
- electrics.resonance: median resonance effect sign in {-1,0,+1}
- polarity.gasteiger_q50: median Gasteiger partial charge
- polarity.gasteiger_iqr: IQR of Gasteiger charge (variability)
- polarity.tpsa_contrib_q50: median per-atom TPSA proxy
- hbond.donor_ratio: fraction flagged as H-bond donors (0–1)
- hbond.acceptor_ratio: fraction flagged as H-bond acceptors (0–1)
- hetero_r1_median: median #heteroatom neighbors (1-hop)
- neighbors_top: top neighbor tokens by PMI; each item has:
    {token: neighbor token id, pmi: log2 P(A,B)/(P(A)P(B)),
     co_occur_ratio: fraction of this token’s atoms with such a neighbor}

EXAMPLE:
{
  "token_id": 112,
  "support_count": 49708,
  "primary_symbol": "C",
  "is_mixed": true,
  "symbol_distribution": {"C": 26016, "O": 23689, "Hg": 3},
  "mixture_entropy": 0.9992911271,
  "env_type": "chain",
  "env_distribution": {"chain": 49689, "ring": 19},
  "aromatic_ratio": 0.0,
  "conjugated_ratio": 0.0001005874,
  "median_degree": 2.0,
  "median_ring_size": 6.0,
  "hybridization": "sp3",
  "electrics": {"inductive": 0, "resonance": 0},
  "polarity": {
    "gasteiger_q50": 0.0339449055,
    "gasteiger_iqr": 0.4922527119,
    "tpsa_contrib_q50": 0.0
  },
  "hbond": {"donor_ratio": 0.0014082240, "acceptor_ratio": 0.4765631287},
  "hetero_r1_median": 0.0,
  "neighbors_top": [
    {"token": 30,  "pmi": 6.3621896439, "co_occur_ratio": 0.9395067192},
    {"token": 53,  "pmi": 3.2824712421, "co_occur_ratio": 0.1208256216},
    {"token": 112, "pmi": 1.9914829505, "co_occur_ratio": 0.0966444033},
    {"token": 58,  "pmi": 1.3526284184, "co_occur_ratio": 0.0445401143},
    {"token": 430, "pmi": 0.5000017521, "co_occur_ratio": 0.0427898930}
  ]
}
\end{chemprompt}

%% file: prompt/input_analogy_example.tex
\begin{chemprompt}{Input Retrieved Analogues Format}
SMILES: Fc1cc(cc(F)c1)CC(NC(=O)C)C(O)C[NH2+]C1(CCCCC1)c1cc(ccc1)C1CCOC1

DTS: A410 A188 A434 A434 A434 A188 A410 A434 A139 A17 A425 A505 A263 A56 A100 A332 A100 A296 A296 A389 A389 A215 A389 A389 A424 A107 A503 A107 A107 A107 A146 A146 A229 A229 A146

Similarity: 0.928

GT: yes

SAF:
[A410, Atom#0, C]: gasteiger_q50=-0.059, gasteiger_iqr=0.145, hba_ratio=0.479, hbd_ratio=0.006, aromatic_ratio=0.515, conjugated_ratio=0.517, ring_ratio=0.000, median_degree=2.000, neighbors_top=[{'token': 124, 'pmi': 6.908255240747004, 'co_occur_ratio': 0.2788412820795242}, {'token': 188, 'pmi': 6.555685343667383, 'co_occur_ratio': 0.5088445864081947}, {'token': 109, 'pmi': 4.684120859168923, 'co_occur_ratio': 0.13898006388368764}, {'token': 410, 'pmi': 4.239188136752083, 'co_occur_ratio': 0.4191651062892389}, {'token': 484, 'pmi': -1.0447276102119363, 'co_occur_ratio': 0.08201343760326027}]  

...

SMF:
{'TPSA': 75.17, 'LogP': 3.299, 'MolWt': 487.611, 'HBA': 3, 'HBD': 3, 'NumAromaticRings': 2, 'NumRotatableBonds': 9, 'NumHeteroatoms': 7, 'FormalCharge': 1} 

\end{chemprompt}

%% file: prompt/input_tested_molecule.tex
\begin{chemprompt}{Input Tested Molecule Format}
SMILES: Fc1cc(cc(F)c1)CC(NC(=O)C)C(O)C[NH2+]C1(CCCCC1)c1cc(ccc1)C1CCOCOC1

DTS: A410 A188 A434 A434 A434 A188 A410 A434 A139 A17 A425 A505 A263 A56 A100 A332 A100 A296 A296 A389 A389 A215 A389 A389 A424 A107 A503 A107 A107 A107 A146 A229 A229 A229 A312 A312 A229

SAF:

[A410, Atom#0, C]: gasteiger_q50=-0.059, gasteiger_iqr=0.145, hba_ratio=0.479, hbd_ratio=0.006, aromatic_ratio=0.515, conjugated_ratio=0.517, ring_ratio=0.000, median_degree=2.000, neighbors_top=[{'token': 124, 'pmi': 6.908255240747004, 'co_occur_ratio': 0.2788412820795242}, {'token': 188, 'pmi': 6.555685343667383, 'co_occur_ratio': 0.5088445864081947}, {'token': 109, 'pmi': 4.684120859168923, 'co_occur_ratio': 0.13898006388368764}, {'token': 410, 'pmi': 4.239188136752083, 'co_occur_ratio': 0.4191651062892389}, {'token': 484, 'pmi': -1.0447276102119363, 'co_occur_ratio': 0.08201343760326027}]

…

SMF:
{'TPSA': 84.400, 'LogP': 3.274, 'MolWt': 517.637, 'HBA': 4.000, 'HBD': 3.000, 'NumAromaticRings': 2.000, 'NumRotatableBonds': 9.000, 'NumHeteroatoms': 8.000, 'FormalCharge': 1.000}
\end{chemprompt}

%% file: prompt/stage1.tex
\begin{chemprompt}{Stage I — Round 1: Prior Schema Analysis}
System:
You are a chemistry and molecular property prediction expert.
Your task is to understand the available features for molecular analysis.

User:
I will provide a list of features. For each, briefly state:
1) what it measures, 2) which property it relates to, 3) when it matters.

Atom-Level (AtomDisc KB): support_count, primary_symbol, is_mixed, mixture_entropy,
gasteiger_q50, gasteiger_iqr, hba_ratio, hbd_ratio, aromatic_ratio, conjugated_ratio,
ring_ratio, median_degree, neighbors_top (PMI)

Molecule-Level (RDKit): TPSA, LogP, MolWt, HBA, HBD,
NumAromaticRings, NumRotatableBonds, NumHeteroatoms, FormalCharge
\end{chemprompt}

\begin{chemprompt}{Stage I — Round 2: Task Understanding}
System:
Continue as a chemistry expert. Focus on the task.

User:
Task: {instruction}

Please analyze:
1) the chemical/biological significance;
2) key molecular characteristics for this task;
3) structural cues separating positive vs. negative samples.
\end{chemprompt}

\begin{chemprompt}{Stage I — Round 3: Feature Selection (JSON Output)}
System:
Select the most relevant features and justify them concisely.

User:
Task: {instruction}

Output JSON:
{
  "atom_features": ["feature1", "feature2", ...],
  "molecule_features": ["feature1", "feature2", ...],
  "feature_descriptions": {
    "feature1": "1-2 sentences on meaning/relevance",
    "feature2": "...",
    "...": "..."
  }
}

IMPORTANT:
1) Use exact names from the lists;
2) Provide descriptions for ALL selected features;
3) Keep descriptions concise (1–2 sentences).
\end{chemprompt}

%% file: prompt/stage3.tex
\begin{chemprompt}{Stage III: Inference With Analogues and Chemical Priors}
System: 
You are a molecular property prediction expert.
Your task is to predict whether a molecule satisfies a given property based on:
1. Selected relevant features
2. Instance-specific analysis from similar training examples
You must provide:
1. A binary prediction (yes/no)
2. A confidence score P(yes) from 0 to 100

User: 
Task: {instruction}

FD: {feature_descriptions}

SMILES: {smiles} 

DTS: {query_tokens}

SAF: {atom_features}

SMF {molecule_features}

ISA: {similar_analysis}

Based on all the information above, please predict whether this molecule satisfies the task.

Your response MUST follow this exact format:

<analysis>
Brief reasoning based on features and similar examples
</analysis>

<answer>yes</answer> or <answer>no</answer>

<confidence>XX</confidence>

Where:
- <answer> MUST be exactly "yes" or "no" (lowercase)
- <confidence> is P(yes) from 0-100 (integer)
- If you answer "yes", confidence should be > 50
- If you answer "no", confidence should be < 50

CRITICAL: You MUST include all three tags (<analysis>, <answer>, <confidence>) in your response.   
\end{chemprompt}

%% file: prompt/DA.tex
\begin{chemprompt}{Directly Answer without Analogues}
System: 
You are an expert medicinal chemist. Directly answer the question with "yes" or "no".

User:
Task: {TASK_NAME}

SMILES: {SMILES}

Your response MUST follow this exact format:
<answer>yes</answer> or <answer>no</answer>
<confidence>XX</confidence>

Where:
- <answer> MUST be exactly "yes" or "no" (lowercase)
- <confidence> is P(yes) from 0-100 (integer)
- If you answer "yes", confidence should be > 50
- If you answer "no", confidence should be < 50

CRITICAL: You MUST include all three tags (<answer>, <confidence>) in your response.   

Answer: 
\end{chemprompt}

%% file: prompt/CoT.tex
\begin{chemprompt}{CoT without Analogues}
System:
You are an expert medicinal chemist. Your task is to analyze a molecule given its SMILES representation and predict whether it satisfies the given condition.

User:
Task: {TASK_NAME}

Instructions:
1. Carefully analyze the SMILES structure
2. Consider molecular properties, functional groups, and structural features
3. Provide clear reasoning based on chemical principles
4. Give your final answer as "yes" or "no"

Format your response as:
- First, provide your analysis and reasoning (2-3 sentences)
- Then, provide your final answer on a new line starting with "Answer: " followed by "yes" or "no"

Question: {QUESTION}
SMILES: {SMILES}

Your response MUST follow this exact format:

<analysis>
Brief reasoning based on features and similar examples
</analysis>
<answer>yes</answer> or <answer>no</answer>
<confidence>XX</confidence>

Where:
- <answer> MUST be exactly "yes" or "no" (lowercase)
- <confidence> is P(yes) from 0-100 (integer)
- If you answer "yes", confidence should be > 50
- If you answer "no", confidence should be < 50

CRITICAL: You MUST include all three tags (<analysis>, <answer>, <confidence>) in your response.      

Your analysis and answer:
\end{chemprompt}

%% file: output/stage1-round1.tex
\begin{chemprompt}{Stage I — Round 3: Feature Selection Output}
{
  "atom_features": [
    "primary_symbol", "gasteiger_q50", "gasteiger_iqr", "hba_ratio", "hbd_ratio", "aromatic_ratio", "conjugated_ratio", "ring_ratio", "median_degree", "neighbors_top"
  ],
  
  "molecule_features": [
    "TPSA", "LogP", "MolWt", "HBA", "HBD", "NumAromaticRings", "NumRotatableBonds", "NumHeteroatoms", "FormalCharge"
  ],

  "feature_descriptions": {
    "primary_symbol": "Element identity (e.g., N, O) indicates potential basic centers and H-bonding atoms needed to engage BACE-1’s catalytic Asp residues.",
    "gasteiger_q50": "Median partial charge reflects local polarity/basicity, informing ionic and H-bond interactions with the Asp dyad.",
    "gasteiger_iqr": "Charge variability signals context sensitivity of electronics, affecting strength and orientation of binding in the active site.",
    "hba_ratio": "Fraction acting as H-bond acceptors; supports complementary interactions in BACE’s polar pocket and conserved waters.",
    "hbd_ratio": "Fraction acting as H-bond donors; enables directional H-bonds but excess donors can reduce CNS permeability.",
    "aromatic_ratio": "Proportion of aromatic character; contributes π-interactions and hydrophobic contacts in BACE subsites while influencing permeability.",
    "conjugated_ratio": "Fraction in conjugated systems; affects electron delocalization and polarizability, tuning binding and lipophilicity.",
    "ring_ratio": "Fraction of atoms in rings; increases rigidity to align pharmacophore and reduce binding entropy costs.",
    "median_degree": "Typical number of bonded neighbors; shapes steric bulk and substitution pattern for pocket fit.",
    "neighbors_top": "Most common neighboring atoms indicate functional group context that drives specific binding interactions.",
    "TPSA": "Topological Polar Surface Area; moderate values balance BBB penetration with sufficient polarity for binding.",
    "LogP": "Lipophilicity; needs to be tuned for CNS permeability while avoiding nonspecific hydrophobic binding.",
    "MolWt": "Molecular weight; size influences BBB penetration and the ability to span BACE subsites without being overly large.",
    "HBA": "Count of hydrogen bond acceptors; facilitates binding but high counts raise polarity and desolvation.",
    "HBD": "Count of hydrogen bond donors; enables key H-bonds yet excessive donors hinder permeability.",
    "NumAromaticRings": "Number of aromatic rings; provides π/hydrophobic contacts and shape, though too many can hurt solubility/permeability.",
    "NumRotatableBonds": "Molecular flexibility; fewer rotatable bonds improve binding efficiency and CNS drug-likeness.",
    "NumHeteroatoms": "Total non-C/H atoms; correlates with polarity and specific interactions, with high counts potentially impairing BBB crossing.",
    "FormalCharge": "Net formal charge; 0 to +1 is typically favored for BBB penetration and ionic interaction with catalytic Asp residues."
  }
}
\end{chemprompt}

%% file: output/example_output_classification_bace.tex
\begin{chemprompt}{Example Output for Classification Task -- BACE}
SMILES: "Fc1cc(cc(F)c1)CC(NC(=O)C)C(O)C[NH2+]C1(CCC(=O)NC1)c1cc(ccc1)C(C)(C)C"

GT: yes

Response:
<analysis> The query molecule has very similar structural features to the top positive examples, particularly the first positive example which shares nearly identical molecular features (HBD=4, TPSA=95.04, LogP=2.04, MolWt=488.60, NumAromaticRings=2). The key functional atoms including the amide carbonyl (A263), hydroxyl group (A332), and ammonium group (A397) provide essential hydrogen bonding interactions with BACE-1 catalytic aspartates. The aromatic rings (A410, A188, A434 patterns) enable \u03c0-stacking with Tyr71. The TPSA (95.04) falls within the optimal range for CNS penetration, and LogP (2.04) is in the preferred range for brain exposure. The most similar negative example differs by having a ketone instead of an amide in the ring system, suggesting the amide group in the query molecule provides better binding interactions. The high structural similarity to multiple positive examples and optimal CNS drug-like properties strongly support BACE-1 inhibition. </analysis>

<answer>yes</answer>

<confidence>85</confidence>
\end{chemprompt}

%% file: output/example_output_classification_bbbp.tex
\begin{chemprompt}{Example Output for Classification Task -- BBBP}
SMILES: "C3=C(C2(C1=CC=CC=C1)C(NCN2)=O)C=CC=C3"

GT: yes

Response:
<analysis>The query molecule is small (MolWt 238) and neutral, with moderate lipophilicity (LogP 1.61) and low polarity (TPSA 41.1). TPSA < 60 is typically favorable for BBB penetration. Although HBD = 2 can be somewhat limiting, several top positive neighbors have similar HBD (2) and comparable LogP/TPSA ranges (e.g., TPSA ~58, LogP ~0.5\u20131.77), and share key functional motifs (amide carbonyl [A482] with adjacent nitrogens [A154/A151], and aromatic rings). The nearest positive examples have notably higher similarity (0.36\u20130.50) than the negatives (\u22640.21), supporting a positive prediction. Overall, the feature profile and neighbor evidence favor BBB penetration.</analysis>

<answer>yes</answer>

<confidence>74</confidence>
\end{chemprompt}

%% file: output/example_output_classification_clintox.tex
\begin{chemprompt}{Example Output for Classification Task -- Clintox}
SMILES: "CC1=C(C(C(=C(N1)C)C(=O)OCC[NH+](C)Cc2ccccc2)c3cccc(c3)[N+](=O)[O-])C(=O)OC",

GT: no

Response: 
"<analysis> The query shares the same nitro\u2011aromatic core and ester functionalities as multiple highly similar NEGATIVE examples (similarities 0.621, 0.446, 0.444), with very close atom-level features (A113/A20 aromatic carbons, A340/A302 oxygens, nitro group A469/A200). Its molecule-level descriptors (TPSA 112, LogP 2.26, HBA 7, HBD 2, MW 480) align closely with these negatives (TPSA ~108\u2013117, LogP ~2.18\u20133.20, HBA 7\u20138). The POSITIVE set is less similar (max ~0.205) and shows more varied scaffolds and properties. Although the query\u2019s formal positive charge (1) may modulate some properties, the nearest-neighbor pattern strongly favors the negative class.</analysis>

<answer>no</answer>

<confidence>34</confidence>
\end{chemprompt}

%% file: output/example_output_classification_hiv.tex
\begin{chemprompt}{Example Output for Classification Task -- HIV}
SMILES: "Cc1cn(C(CO)OC(CO)CN(O)C2C3CC4CC(C3)CC2C4)c(=O)[nH]c1=O"
GT: no
Response:
<analysis> The query molecule shares a highly similar core structure with both positive and negative examples, featuring the characteristic uracil-like ring system (A457, A457, A154, A416, A412, A482 pattern) that appears in anti-HIV compounds. However, several concerning features suggest it may not exhibit anti-HIV activity: 1. Molecular weight (409.483) is significantly higher than most positive examples (224-314 range) and approaches the upper limit for drug-like molecules, potentially hindering binding pocket accommodation. 2. High rotatable bond count (8) indicates excessive flexibility, which typically reduces binding affinity to rigid enzyme pockets - this is higher than most positive examples (2-6 range). 3. The complex bicyclic substituent introduces substantial steric bulk that may not fit well into known anti-HIV enzyme binding sites. 4. While TPSA (128.02) and HBA/HBD counts are reasonable for anti-HIV agents, the most similar negative example (similarity 0.529) shares nearly identical molecular features (TPSA=128.02, HBA=8, HBD=4) and also contains the problematic high rotatable bond count. The structural similarity to negative examples, particularly the high molecular weight and flexibility, outweighs the shared core with positive compounds.</analysis>
<answer>no</answer>
<confidence>35</confidence>
\end{chemprompt}

%% file: output/example_output_classification_tox21.tex
\begin{chemprompt}{Example Output for Classification Task -- Tox21}
SMILES: "COc1ccc2c3c1O[C@H]1[C@@H](O)C=C[C@H]4[C@@H](C2)NCC[C@]314"
GT: no
Response: 
<analysis>  The query molecule has key features that align with known ER-LBD activators: HBD=2 (including a phenolic -OH group at A423 with hbd_ratio=0.998), HBA=4, moderate LogP=1.16, low TPSA=50.72, and rigid structure (NumRotatableBonds=1). However, it only has one aromatic ring (NumAromaticRings=1), which is suboptimal compared to typical ER agonists that have 2-4 aromatic rings for optimal hydrophobic contact. The most similar negative example (similarity=0.448) shares the same steroid-like scaffold with only one aromatic ring and was inactive, while positive examples with higher similarity tend to have more aromatic character. The presence of the essential phenolic -OH group is favorable, but the limited aromatic system likely reduces binding efficiency to the hydrophobic ER pocket.</analysis>
<answer>no</answer>
<confidence>35</confidence>",
\end{chemprompt}

%% file: output/example_output_classification_sider.tex
\begin{chemprompt}{Example Output for Classification Task -- Sider}
SMILES: "CC1=C(C2=CC3=NC(=CC4=C(C(=C(N4)C=C5[C@@]6([C@@H](C(=CC=C6C(=N5)C=C1N2)C(=O)OC)C(=O)OC)C)C)CCC(=O)OC)C(=C3C)CCC(=O)O)C=C"
GT: yes
Response:
<analysis> The query molecule shows several concerning features for carcinogenicity. It has a high aromatic ratio (0.595-0.939 across atoms) and extensive conjugation (0.951-1.000), which enables DNA intercalation. The presence of 3 aromatic rings and polycyclic aromatic structure is a classic structural alert for carcinogenicity. The high LogP (6.719) indicates significant lipophilicity for bioaccumulation, while the molecular weight (718.807) suggests potential bioavailability issues, though the molecule may still reach nuclear targets. The similar positive examples share key structural motifs - particularly the polycyclic aromatic systems with nitrogen heteroatoms and carboxylic acid/carboxylate groups. The top positive example has very similar atom-level features (same token patterns A113, A160, A162) and comparable molecular properties (TPSA 152.19 vs 173.56, similar aromatic ring count). The negative examples tend to have simpler aromatic systems, lower molecular complexity, and different functional group patterns. The combination of DNA-intercalating aromatic systems, metabolic activation potential from heteroatoms, and structural similarity to known carcinogens supports a positive prediction.</analysis>
<answer>yes</answer>
<confidence>78</confidence>
\end{chemprompt}

%% file: results/regression.tex
\begin{table*}[htbp]
\centering
\scriptsize
\setlength{\tabcolsep}{6pt}
\renewcommand{\arraystretch}{1.15}
\caption{Regression benchmarks (RMSE~$\downarrow$) across six datasets (Aqsol, Caco2, ESOL, FreeSolv, LD50, Lipophilicity). Entries marked ``---'' were not reported. The \emph{Average} column is computed per method over the available datasets. The best results are shown in \textbf{bold}, and the runner-up results are \underline{underlined}.}
\label{tab:rmse_summary}
\resizebox{\textwidth}{!}{%
\begin{tabular}{lccccccc}
\toprule
\textbf{Method} & \textbf{Aqsl} & \textbf{Caco2} & \textbf{ESOL} & \textbf{FreeSolv} & \textbf{LD50} & \textbf{Lipophilicity} & \textbf{Average} \\
\midrule
CoT          & 2.8430 & 13.7336 & 0.8944 & 3.4312   & 737.9134 & 1.2098 & 126.6709 \\
CoT+FS       & 1.3486 & 8.6642  & 0.9134 & 3.2863   & 12.7961  & 1.2479 & 4.7094 \\
DA           & 1.4254 & 17.9927 & 1.0194 & 2.4534   & 297.4678 & 1.3517 & 53.6184 \\
DA+FS        & 1.4581 & 0.6335  & 1.0497 & 2.5510   & 1.1385   & 1.4555 & 1.3810 \\
MolRAG       & ---    & ---     & 3.2806 & 6.0827   & ---      & 1.1251 & --- \\
LLM4SD       & ---    & ---     & 0.5200 & 2.6200   & ---      & \underline{}{0.6800} & --- \\
TokenMol     & \textbf{0.7350} & 0.6110  & \textbf{0.5930} & \underline{1.2250}   & 0.9400   & \textbf{0.6450} & \underline{0.7915} \\
MoleculeSTM  & 1.3293 & \underline{0.5409}  & 1.1770 & 1.2880   & \underline{0.8047}   & 0.6944 & 0.9724 \\
\midrule
ChemATP      & \underline{1.1266} & \textbf{0.5041} & \underline{0.6504} & \textbf{1.0177} & \textbf{0.7095} & {0.7143} & \textbf{0.7871} \\
\bottomrule
\end{tabular}%
}
\end{table*}

%% file: results/knn.tex
\begin{table*}[htbp]
\centering
\scriptsize
\setlength{\tabcolsep}{6pt}
\renewcommand{\arraystretch}{1.15}
\caption{Regression benchmarks (RMSE~$\downarrow$) across six datasets (Aqsol, Caco2, ESOL, FreeSolv, LD50, Lipophilicity). The \emph{Average} column is computed per method over all six datasets. To demonstrate the effectiveness of the reasoning capability of LLMs, we further compare the KNN method with our ChemATP on regression tasks. As shown in Table~\ref{tab:rmse_knn_summary}, our ChemATP greatly outperform than the KNN or weighted-KNN methods. 
Importantly, these KNN baselines use {exactly the same retrieval setup} as ChemATP (the same similarity metric and the same top-$k$ neighbors), and differ only in the prediction head: KNN directly averages the retrieved labels, while weighted-KNN performs similarity-weighted averaging.
Therefore, the substantial gap between ChemATP and the two nearest-neighbor baselines indicates that the improvements cannot be explained by nearest-neighbor signals alone (i.e., simply copying/aggregating retrieved labels or priors). 
Instead, the LLM effectively {integrates} the retrieved evidence with task instructions and chemical context to perform non-trivial reasoning beyond local interpolation, which leads to consistently better performance across all benchmarks.}
\label{tab:rmse_knn_summary}
\resizebox{\textwidth}{!}{%
\begin{tabular}{lccccccc}
\toprule
\textbf{Method} & \textbf{Aqsol} & \textbf{Caco2} & \textbf{ESOL} & \textbf{FreeSolv} & \textbf{LD50} & \textbf{Lipophilicity} & \textbf{Average} \\
\midrule
KNN           & {1.6771} & {0.5779} & {1.5175} & {10.1398} & {0.8169} & {0.9199} & {2.6082} \\
Weighted-KNN  & {1.6580}    & {0.5623}    & {1.5218} & {9.6172}  & {0.8054}    & {0.9067}    & {2.5119} \\
\midrule
ChemATP      & {1.1266} & {0.5041} & {0.6504} & {1.0177} & {0.7095} & {0.7143} & {0.7871} \\
\bottomrule
\end{tabular}%
}
\end{table*}